\documentclass[10pt,twocolumn,letterpaper]{article}

\usepackage{iccv}              % To produce the CAMERA-READY version
\usepackage{graphicx}
\usepackage{booktabs}
\usepackage{xcolor}
\usepackage{multirow}
\usepackage{amsmath,amssymb,amsfonts}
\usepackage{mathrsfs}
\usepackage[title]{appendix}
\usepackage{textcomp}
\usepackage{manyfoot}
\usepackage{algorithm}
\usepackage{algpseudocode}
\usepackage{listings}
\usepackage{pifont}
\usepackage{soul}
\usepackage{enumitem}
\usepackage{tabularx}
\usepackage{lipsum}
\usepackage{mathbbol}
\usepackage{tabularray}
\usepackage[utf8]{inputenc}
\usepackage{color, colortbl}
\usepackage{wrapfig}

\definecolor{deepblue}{rgb}{0,0,0.5}

\definecolor{iccvblue}{rgb}{0.21,0.49,0.74}
\usepackage[pagebackref,breaklinks,colorlinks,allcolors=iccvblue]{hyperref}

\usepackage[capitalize]{cleveref}
\crefname{section}{Section}{Sections}
\crefname{table}{Table}{Tables}
\crefname{figure}{Figure}{Figure}
\newcommand{\model}{\texttt{DualFit}\xspace}

\def\paperID{*****} % *** Enter the Paper ID here
\def\confName{ICCV}
\def\confYear{2025}

\title{DualFit: A Two-Stage Virtual Try-On via Warping and Synthesis}

\author{
Minh Tran$^1$,
    Johnmark Clements$^1$,
    Annie Prasanna$^1$,
    Tri Nguyen$^2$, Ngan Le$^1$ \\ 
    $^1$University of Arkansas, $^2$Coupang, Inc. \\
    {\small \url{https://uark-aicv.github.io/DualFit}}
}

\begin{document}
% \maketitle

\twocolumn[{
\maketitle
\begin{center}
   \includegraphics[width=1.0\textwidth]{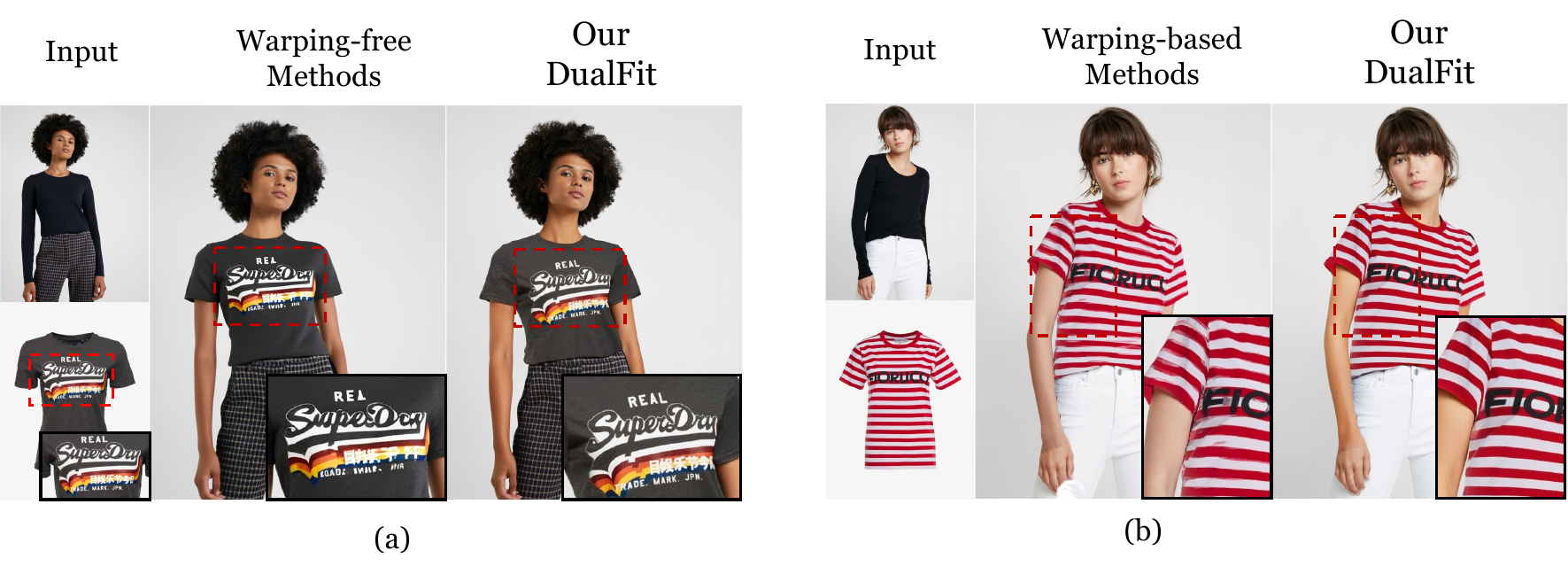}
   %\captionof{figure}{Current warping-free methods tend to omit or blur critical garment details, such as printed text. \hl{Update with both warping-based and warping-free methods, highlighting the limitatons of each}}
   \captionof{figure}{Comparison of warping-free, warping-based, and our proposed DualFit methods on VTON results. Left: Warping-free methods produce smooth and perceptually realistic outputs but often fail to preserve fine-grained garment details such as graphics text in this case. Right: Warping-based methods better retain garment textures but frequently introduce misalignment artifacts and unnatural seams.}
\label{fig:teaser}
\end{center}
}]

\begin{abstract}
Virtual Try-On (VTON) technology has garnered significant attention for its potential to transform the online fashion retail experience by allowing users to visualize how garments would look on them without physical trials. While recent advances in diffusion-based warping-free methods have improved perceptual quality, they often fail to preserve fine-grained garment details such as logos and printed text—elements that are critical for brand integrity and customer trust. In this work, we propose \model, a hybrid VTON pipeline that addresses this limitation by two-stage approach. In the first stage, \model warps the target garment to align with the person image using a learned flow field, ensuring high-fidelity preservation. In the second stage, a fidelity-preserving try-on module synthesizes the final output by blending the warped garment with preserved human regions. Particularly, to guide this process, we introduce a preserved-region input and an inpainting mask, enabling the model to retain key areas and regenerate only where necessary, particularly around garment seams. Extensive qualitative results show that \model achieves visually seamless try-on results while faithfully maintaining high-frequency garment details, striking an effective balance between reconstruction accuracy and perceptual realism.
%—particularly in warping-free methods leveraging latent diffusion models—have led to improved perceptual quality, they often struggle to preserve fine-grained visual details such as garment patterns, logos, and printed text. In this work, we propose a novel VTON pipeline that revisits warping-based techniques to explicitly address this shortcoming. ...
\end{abstract}    
\section{Introduction}
\label{sec:intro}

Image-based Virtual Try-On (VTON) has gained significant popularity in recent years due to its potential impact on the online shopping industry~\cite{islam2024deep, jiang2024fitdit, xie2023gp, choi2024improving, chong2024catvton, wang2025mv, chong2025catv2ton, luan2025mc, wei2025dh, shen2025mfp, luo2025crossvton} %due to the applications it could have on the online shopping industry. 
This technology would allow users to virtually try on clothing by creating a realistic generated image of themselves wearing a different clothing item. By allowing customers to see how garments might look before making a purchase, VTON can reduce the uncertainty often associated with online shopping. As a result,  it helps to increased customer satisfaction and fewer returns, ultimately saving retailers time and resources. Despite its promise, current VTON models still face significant challenges in achieving fully realistic results. Two key challenges are reconstruction accuracy and perceptual quality.
Reconstruction refers to the model's ability to recreate the ground truth image in the generated image. This includes preserving fine details such as patterns, printed text, or logos. It reflects how well the model maintains the fidelity of both the human figure and the garment. Perceptual quality, on the other hand, refers to the model’s ability to generate an image that looks natural and is visually appealing to look, which is often referred to as realism. Both aspects are critical for building user trust and providing a seamless virtual fitting experience.

% \begin{figure*}
%     \centering
%     \includegraphics[width=1\linewidth]{Figs/teaser.pdf}
%     \caption{}
%     \label{fig:teaser}
% \end{figure*}

Recent literature has seen a growing interest in \emph{warping-free} methods that leverage advances in latent diffusion models~\cite{rombach2022high}. These approaches typically remove the original garment from the person image and synthesize a new try-on image in an end-to-end manner, using the target garment as a conditioning input to the diffusion model~\cite{choi2024improving, chong2024catvton, morelli2023ladi}. While warping-free methods produce visually smooth results with high perceptual quality, they often fail to they often fail to achieve high-fidelity reconstruction, particularly in preserving fine-grained visual details. This limitation stems from the nature of latent-space representations, where high-frequency information such as logos, printed text, and intricate patterns is often lost during the encoding process in both the variational autoencoder (VAE) and the denoising diffusion model. As illustrated in Figure~\ref{fig:teaser}, current warping-free methods tend to omit or blur critical garment details, such as printed text. From the perspective of fashion retailers and brands, faithfully reconstructing garment patterns, prints, and logos is critical because these elements embody the brand’s identity, craftsmanship, and value proposition.

Motivated by this challenge, in this work, we propose \model, a VTON pipeline that specifically addresses the limitations of existing warping-free methods. \model operates in two stages: it first performs warping to align the target garment with the body in the person image, and then synthesizes the final try-on result by blending the warped garment with the individual's appearance. Unlike warping-free approaches, which generate garments directly from the model, our \model predicts a flow field to transform the original garment. This allows for the preservation of visual fidelity, including critical fine-grained details such as logos, graphics, and printed text.

% \hl{Those two paragraphs should be rewritten. The previous paragraph can introduce DualFit, whereas this one can highlight the difference between them.}

\begin{figure*}
    \centering
    \includegraphics[width=1\linewidth]{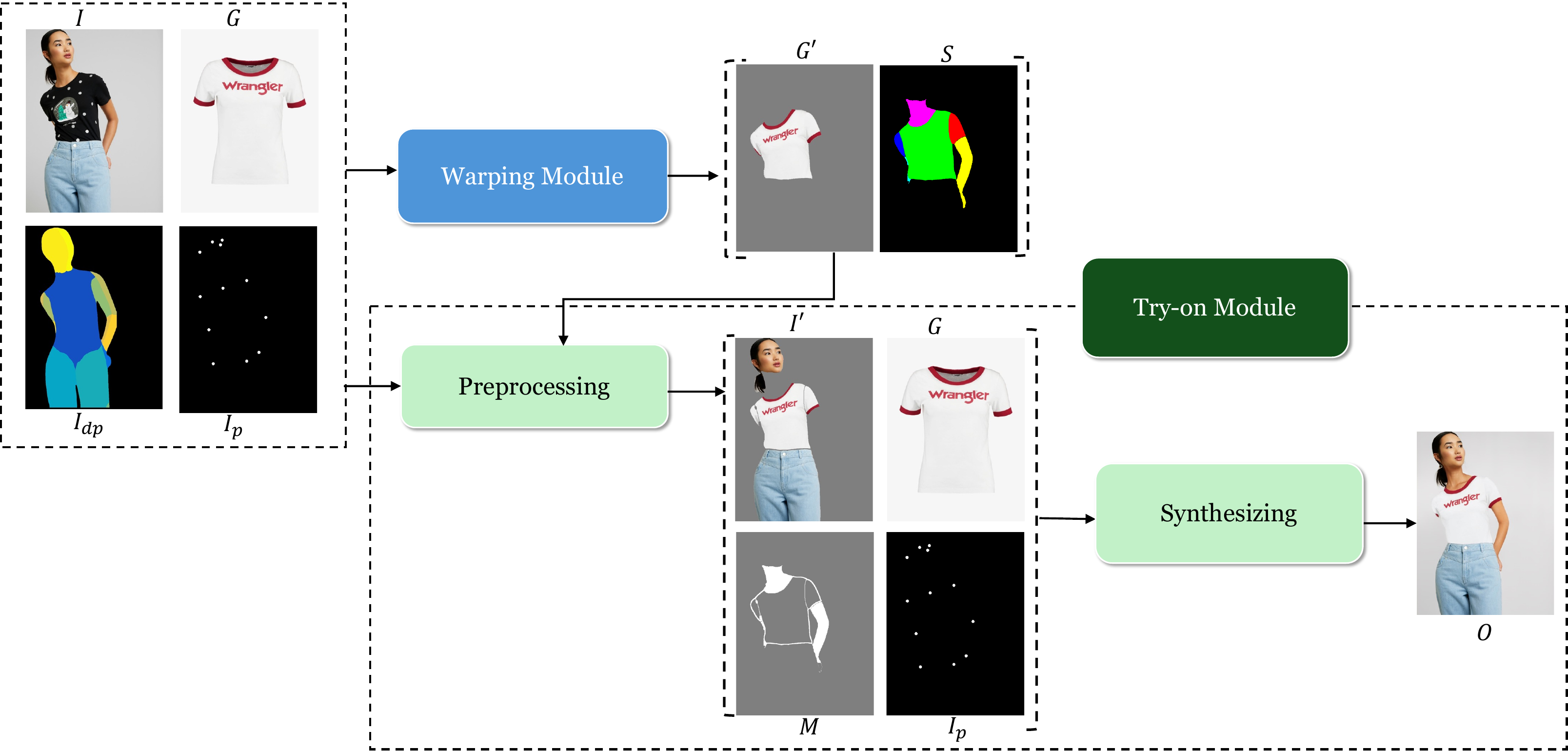}
    \caption{Overview of our image-based virtual try-on pipeline. The pipeline consists of two stages: (1) \textbf{Warping Stage}, where the Warping Module takes the person image $I$, in-shop garment $G$, human densepose map $I_{dp}$, and pose heatmap $I_p$ to predict the warped garment $G'$ aligned to the target body shape, along with the upper body parsing map $S$; and (2) \textbf{Try-on Stage}, where the Try-on Module uses the warped garment $G'$, the preserved-region input $I'$, inpaint mask $M$, flat garment $G$, and pose $I_p$ to synthesize the final try-on result $O$ that seamlessly blends the garment onto the person.}
    \label{fig:pipeline}
\end{figure*}

Among existing VTON methods, GPVTON~\cite{xie2023gp} is the most closely related to our approach. In GPVTON~\cite{xie2023gp}, the second stage of synthesis is typically carried out using generative adversarial networks (GANs) to render the final image by combining the warped garment and preserved human parts. This rendered output however often fails to maintain high-frequency details, as the generation process inherently reduces resolution. To mitigate this, they overlay the warped garment onto the GAN-rendered output as a post-processing step. While this preserves garment fidelity, it often results in unnatural boundaries between the garment and the human body, producing synthetic-looking outputs.
In contrast, our proposed \model pipeline integrates a fidelity-preserving try-on module that produces final outputs with both smooth perceptual quality and high-detail preservation. Specifically, we design a preserved-region input and an inpainting mask to guide the model on which areas to retain and which to regenerate. To obtain the preserved region, we remove the entire upper body (excluding the head and hair) based on the predicted human parsing map. We then overlay the warped garment onto this upper-body-removed image. Additionally, we remove the boundaries between different garment sections (left sleeve, torso, right sleeve), which often contain artifacts due to the assembly of warped parts. By allowing the try-on module to generate these boundary regions, we achieve smoother try-on results. The removed regions from the preserved image form the inpainting mask, which directs the model where new content needs to be generated. Examples of the preserved region image $I'$ and the inpainting mask $M$ are shown in~\cref{fig:pipeline}. To further enhance generation quality, we also condition the try-on process on the flat input garment. As demonstrated in Figure~\ref{fig:teaser}, our approach yields visually seamless try-on results while faithfully preserving fine details such as text, logos, and graphics.

\section{Related Work}
The existing image-based
VTON methods can be divided into warping-based and warping-free approaches.

\textbf{Warping-based }methodologies follow a two-stage pipeline ~\cite{han2018viton, wang2018toward, yu2019vtnfp, issenhuth2020not, chopra2021zflow, choi2021viton, lee2022high, fele2022c, yang2022full, gou2023taming, chen2023size, xie2023gp, yan2023linking}. In the first stage, the garment is warped to match the person’s pose and body. During the second stage, the garment is combined with the image of the person to generate the final try-on result. There are various techniques that have been proposed for the warping process. The process includes Thin Plate Spline (TPS) transformations ~\cite{duchon1977splines, han2018viton, minar2020cp, lee2019viton, yang2020towards}, flow-based warping ~\cite{zhou2016view, bai2022single, chopra2021zflow, ge2021parser, han2019clothflow, he2022style}, and landmark-based alignment ~\cite{liu2021toward, xie2020lg, yan2023linking, chen2023size}. The widely used methods are TPS based methods which are used in earlier VTON works to deform the garment smoothly. Other approaches include Flow based approaches where it predicts dense flow fields to warp garment pixels. Landmark guided methods use body and clothing landmarks for alignment.
For the image synthesis phase, there are some methods which improve fidelity by using human parsing maps~\cite{yu2019vtnfp,yang2020towards, choi2021viton} as additional resource. Whereas other methods improve the generative architecture ~\cite{choi2021viton,dong2020fashion,fele2022c} by modifying normalization layers or integrating advanced modules. Recently, the use of diffusion models~\cite{rombach2022high} has gained popularity over GAN based generators~\cite{goodfellow2020generative}. This is because of the superior image generation capabilities~\cite{gou2023taming, tang2023make} provided by GAN based generators. Such models have improved the visual realism of try-on results. However, warping based methods are still prone to introducing artifacts during garment alignment, plus, they often result in unnatural boundaries between the garment and the human body, producing synthetic-looking outputs.

In contrast, \textbf{warping-free} methods~\cite{baldrati2023multimodal, zhu2023tryondiffusion, morelli2023ladi} which are based on latent diffusion models~\cite{ho2020denoising, rombach2022high}, avoid the warping step to remove deformation artifacts. Interestingly, TryOnDiffusion \cite{zhu2023tryondiffusion} proposed a dual U-Net architecture that showcased the potential of diffusion-based virtual try-on, but it relied on extensive datasets containing image pairs of the same person in diverse poses. This data requirement has motivated recent efforts to adopt large pre-trained diffusion models \cite{ramesh2021zero, rombach2022high, saharia2022photorealistic} instead.
To adapt these models for the try-on task, various strategies have been explored: encoding garments as pseudo-words as in LaDI-VTON \cite{morelli2023ladi}, integrating warping networks like in DCI-VTON \cite{gou2023taming}, altering attention mechanisms in StableVITON \cite{kim2024stableviton} and IDM-VTON \cite{choi2024improving}, applying ControlNet-based garment guidance and GAN sampling in CAT-DM \cite{zeng2024cat}, or simply concatenating masked person and garment images to directly transfer textures as done in TPD \cite{yang2024texture}.
Nevertheless, these approaches inherit key drawbacks from large pre-trained U-Nets: their substantial parameter counts lead to heavy memory consumption and slow inference, limiting practical deployment. Plus, they often fail to they often fail to achieve high-fidelity reconstruction, particularly in preserving fine-grained visual details. This limitation stems from the nature of latent-space representations, where high-frequency information such as logos, printed text, and intricate patterns is often lost during the encoding process in both the variational autoencoder (VAE) and the denoising diffusion model.

In this work, we aim to bridge the gap between fidelity and perceptual realism in VTON. Warping-free methods often sacrifice fine details for smoother appearances, while warping-based approaches tend to introduce artifacts during garment alignment, resulting in unnatural boundaries between the garment and the human body that make outputs appear synthetic. To address these issues, we propose \model, which combines a warping-based alignment step with a fidelity-preserving try-on module. Our method produces final outputs that achieve both high-detail preservation and smooth, realistic perceptual quality.

\section{Method}
Image-based VTON algorithm aims to seamlessly transfer an in-shop garment $G$ onto a specific person $I$ and generate it as the try-on image $O$. Overall, our inference pipeline can be seen in~\cref{fig:pipeline}. This pipeline has two stages: {(1) Warping Stage}, where the Warping Module takes the person image $I$, the in-shop garment $G$, the person’s human densepose map  $I_{dp}$, and the pose heatmap $I_p$ to predict the warped garment $G'$ aligned to the target body shape, along with the upper body parsing map $S$; and {(2) Try-on Stage}, where after preprocessing, the Try-on Module uses the warped garment $G'$, the preserved-region input $I'$, an inpaint mask $M$, the flat input garment $G$ and pose $I_p$ to synthesize the final try-on result $O$ that realistically blends the garment onto the person.

% \subsection{Warping Module}\label{subsection:warping_module}
% We leverage ~\cite{xie2023gp} for their warping module. In fact

\subsection{Warping Module}
\label{subsection:warping_module}
We leverage the Local-Flow Global-Parsing (LFGP) warping module from GP-VTON~\cite{xie2023gp} as our garment warping strategy. 

In general, The LFGP module employs a cascaded flow estimation pipeline using multi-scale pyramid features extracted separately from the person and garment inputs. It estimates local flows for distinct garment regions, such as the left sleeve, right sleeve, and torso, to handle diverse deformations and reduce artifacts that typically occur when a single global flow is applied to the entire garment. Additionally, it predicts an upper-body parsing map $S$ comprising the left hand, right hand, left garment, right garment, torso, and neck. The local flows enable precise warping of individual parts, which are then assembled into a complete warped garment $G'$ using a global garment parsing map. This global parsing map distinguishes three key garment regions: left garment, right garment, and torso.

In details, the module first employs Feature Pyramid Networks (FPNs) to extract multi-scale features of the person (using pose and densepose) and the intact garment (with parsing map). Then, it cascades multiple LFGP blocks to estimate local flows at each scale, where each block refines flow predictions for three garment parts individually. Concurrently, a global garment parsing is estimated to guide the assembly of warped parts. By assigning each pixel in the final warped garment to a specific warped part based on the global parsing, the approach effectively resolves overlap artifacts between parts. This design enables semantically correct and visually consistent garment warping across varying poses and garment shapes, forming a crucial component of our try-on pipeline. To improve training stability for various wearing styles, we also adopt the Dynamic Gradient Truncation (DGT) strategy from GP-VTON~\cite{xie2023gp}. Previous methods apply a fixed gradient truncation mask to preserve certain garment regions, but this can lead to artifacts like texture squeezing or stretching depending on whether the garment is tucked-in or tucked-out in the input person image. DGT addresses this by dynamically deciding whether to truncate gradients based on the wearing style, quantified by the height-width ratio disparity between the flat garment and the warped garment’s torso region. If the ratio indicates tucking-in, gradients inside the preserved region are truncated to avoid forcing the garment to align with the preserved boundary. Conversely, for tucking-out cases, gradients are propagated to penalize misalignment. This adaptive strategy allows the warping module to better generalize across diverse dressing styles.

\subsection{Try-on Module}
\label{sec:tryon_module}
After the warping module, we propose a fidelity-preserving try-on module that generates final outputs with both smooth perceptual quality and high-detail preservation. 

\textbf{Preprocessing. }We first prepare the necessary inputs for the try-on module. Specifically, we construct a preserved-region input $I'$ and an inpainting mask $M$ to guide the model on which areas to retain and which to regenerate. To obtain the preserved region, we remove the entire upper body, excluding the head and hair, based on the predicted upper-body parsing map $S$ from the previous stage, and overlay the warped garment $G'$ onto this upper-body-removed image to create $I'$.

Because the warped garment $G$ often contains artifacts near the boundaries of its warped sections, we remove the borders between different garment regions, specifically the left sleeve, torso, and right sleeve, to allow the try-on module to regenerate these areas, resulting in smoother try-on outputs.

From the upper-body parsing map $S$, we extract three binary masks corresponding to different garment regions: the left sleeve mask $S_l$, the right sleeve mask $S_r$, and the torso mask $S_t$. For each mask $S_*$ (where $ * \in \{l, r, t\}$), we compute a narrow band along its boundary by first applying a morphological erosion operation and then subtracting the eroded mask from the original. The erosion operation $\mathrm{Erode}(S_*, K)$ is defined as
\begin{equation}
\mathrm{Erode}(S_*, K)(x, y) = \min_{(i, j) \in K} S_*(x+i, y+j),
\end{equation}
where $K$ is the structuring element (kernel) defining the local neighborhood. We use a $3 \times 3$ square kernel and perform $n$ iterations of erosion, where $n$ is a configurable parameter controlling the thickness of the narrow band. The band $B_*$ for each mask is computed as
\begin{equation}
B_* = S_* - \mathrm{Erode}(S_*, K)^{(n)},
\end{equation}
where $\mathrm{Erode}(S_*, K)^{(n)}$ denotes applying the erosion operation $n$ times. In our experiments, we use $n=5$ to obtain a narrow band approximately five pixels wide.

By allowing the try-on module to generate these boundary regions, we achieve smoother and more realistic results. To further facilitate this process, we condition the model on the flat garment $G$, providing appearance cues for synthesizing the missing regions.
To construct the complete inpainting mask, we combine the body parts from the human parsing map $S$, including the left hand $S_{e}$, right hand $S_{i}$, and neck $S_{n}$, with the narrow bands $B_l$, $B_r$, and $B_t$, resulting in the final inpainting mask $M$ that guides the try-on model in generating coherent and realistic outputs.
\begin{equation}
    M = S_{e} \cap S_{i} \cap S_n \cap B_l \cap B_r \cap B_t
\end{equation}

\textbf{Synthesizing. }Given the preprocessed input, we employ a Res-UNet-based~\cite{ronneberger2015u} generator $\mathcal{G}$ to synthesize the final try-on output. Specifically, the network is designed as a U-Net architecture augmented with residual connections, enabling effective feature propagation and stable gradient flow during training. The generator progressively downsamples the input through multiple encoder layers and reconstructs the output via corresponding decoder layers, connected through skip connections to preserve spatial details. Each block within the U-Net is implemented as a residual skip-connection block, where intermediate features are refined by residual learning. Batch normalization is applied for stable training, and optional dropout layers can be inserted to improve generalization.

For the loss function, we utilize the pixel-wise $\ell_1$ loss $\mathcal{L}_1$, the perceptual loss~\cite{johnson2016perceptual} $\mathcal{L}_{per}$, and the adversarial loss $\mathcal{L}_{adv}$ for supervising the try-on result $O$. The pixel-wise $\ell_1$ loss plays an important role in preserving the aligned regions of the garment. During training, we use the ground truth warped garment as input so that it is spatially aligned with the ground truth try-on image, enabling the model to learn to copy the warped garment details directly from the input to the output, thereby preserving the garment.
For the generated parts outside the preserved regions, the combination of $\ell_1$ loss and perceptual loss encourages smooth and realistic synthesis, allowing the newly generated regions to blend seamlessly with the preserved warped garment.
The total generator loss $\mathcal{L}^{gen}$ is defined as follows:
\begin{equation}
    \mathcal{L}^{gen} = \mathcal{L}_1 + \mathcal{L}_{per} + \mathcal{L}_{adv}.
\end{equation}

\section{Experiments}

\begin{table*}
    \centering
    \setlength{\tabcolsep}{4pt}
    \caption{Quantitative comparison with recent SOTA warping-free and warping-based VTON methods. Higher PSNR and SSIM and lower FID, LPIPS, and DIST indicate better performance. The best results for each metric are shown in bold, and second-best results are \underline{underlined}.}
    \begin{tabular}{l|l|c|cc|c|cc|c}
    \toprule
        ~ & \multirow{2}{*}{\textbf{Method}} & \multirow{2}{*}{\textbf{Venue}}  & \multicolumn{2}{c|}{\textbf{Reconstruction}}  & \multicolumn{1}{c|}{\textbf{Perceptual Realism}} & \multicolumn{2}{c|}{\textbf{Hybrid}} & \multirow{2}{*}{\textbf{FPS}$\uparrow$}  \\ \cmidrule{4-8}
        ~ & ~ & ~  & PSNR$\uparrow$ & SSIM$\uparrow$ & FID$\downarrow$ & LPIPS$\downarrow$ & DIST$\downarrow$ & ~ \\ \midrule
    \multirow{7}{*}{\rotatebox{90}{Warp-free}} & Stable-VITON~\cite{kim2024stableviton} & CVPR 24 & 20.162 & 0.888 & 8.233 & 0.073 & 0.110 & 0.067 \\ 
        ~ & TPD~\cite{yang2024texture} & CVPR 24 & 21.342 & 0.900 & 8.540 & 0.070 & 0.109 & 0.083  \\ 
        ~ & IDM-VTON~\cite{choi2024improving} & ECCV 24 & 22.185 & 0.870 & 6.290 & 0.102 & 0.080 & 0.059 \\ 
        ~ & SD-VITON~\cite{shim2024towards} & AAAI 24 & -- & 0.896 & 9.750 & 0.070 & -- & -- \\
        % ~ & StableVITON~\cite{kim2024stableviton} & CVPR 24 & -- & 0.852 & 8.698 & 0.0842 & -- & -- \\
        ~ & FITMI~\cite{samy2025revolutionizing} &  Neu.Comp.\&App. 25 & -- & 0.876 & 9.410 & 0.091 & -- & -- \\
        ~ & PA-VTON~\cite{park2025virtual} & Vis.Com.\&Ima. Rep. 25& -- & 0.877& 9.579& 0.091 & -- & -- \\
        ~ & BooW-VTON~\cite{zhang2025boow} & CVPR 25 &  21.800 & 0.862 &  6.885   &0.108  & -- & -- \\ 
        ~ & CatVTON~\cite{chong2024catvton} & ICLR 25 & 21.583 & 0.870 & \underline{5.425} & {0.087} & 0.095 & 0.067 \\ 
        ~ & OOTDiffusion~\cite{xu2025ootdiffusion}& AAAI 25 & -- & 0.878 & 8.810 & 0.071 & -- & -- \\ 
%        ~ CATVTON~\cite{chong2024catvton} & -- & 0.8704 & 5.425 & 0.0565 & 0.0565 & -- & 0.388 \\
         \midrule
        \multirow{3}{*}{\rotatebox{90}{Warp-based}} & GP-VTON~\cite{xie2023gp} & CVPR 23 & \underline{23.009} & 0.894 & 9.197 & 0.080 & \underline{0.077} & \underline{0.567} \\ 
        ~ & LaDI-VTON~\cite{morelli2023ladi} & ACMMM 23 & 21.546 & 0.876 & 6.660 & 0.091 & 0.108 & 0.091 \\ 
        ~ & FIP-VTON~\cite{dam2024time} & ECCV 24 & - & \underline{0.909} & 8.430 & \underline{0.067} & - & 0.495 \\ \cmidrule{2-9}
        ~ & \textbf{\model (ours)} & - & \textbf{24.856} & \textbf{0.913} & \textbf{5.261} & \textbf{0.064} & \textbf{0.046} & \textbf{0.567} \\ \bottomrule
    \end{tabular}
    \label{tab:quan_results}
\end{table*}

\subsection{Experimental Setup}
\textbf{Dataset. }We train and evaluate our method on the VITON-HD dataset~\cite{choi2021viton}. The dataset consists of 11,647 training samples and 2,032 testing samples, all featuring upper-body garments. Each sample includes a model image, which shows a person wearing the garment (captured as a two-thirds body shot from the upper body) with a resolution of $768 \times 1024$ pixels, and a garment image, which depicts the corresponding upper-body garment laid out on a white background, also at a resolution of $768 \times 1024$ pixels.\\

\noindent
\textbf{Metrics. }
We evaluated VTON model performance using standard measurement such as  PSNR~\cite{gonzalez2009digital}, LPIPS~\cite{zhang2018unreasonable}, SSIM~\cite{wang2004image}, FID~\cite{heusel2017gans}, and DISTS~\cite{ding2020image} metrics.
PSNR measures pixel-level reconstruction fidelity but overly penalizes color changes, making it more suitable for assessing reconstruction accuracy than perceptual similarity. 
LPIPS uses deep feature comparisons from a pre-trained network (AlexNet) to better capture perceptual differences; it provides balanced sensitivity to both structural and semantic changes, aligning more closely with human judgment. 
SSIM evaluates luminance, contrast, and structural similarities, favoring structural consistency but also considering color to some extent, making it helpful for measuring reconstruction quality, especially when assessing warping accuracy.
FID computes the Fréchet distance between feature distributions from the Inception network, quantifying overall distributional similarity between generated and real images; though better suited for evaluating large datasets, it reflects both reconstruction and perceptual qualities. Finally, DISTS leverages a pre-trained VGG network to combine structural and textural similarities into a single score, providing a balanced measure of both reconstruction and perceptual quality. Together, these metrics offer complementary perspectives on VTON performance, covering both low-level pixel fidelity and high-level perceptual similarity.\\

These metrics can be grouped into three categories based on their focus: reconstruction metrics (PSNR, SSIM) emphasize pixel-level accuracy; perceptual realism metrics (FID) assess naturalness and visual believability; and hybrid metrics (LPIPS, DISTS) capture a balance between structural fidelity and perceptual similarity. Together, this comprehensive set of metrics allows us to comprehensively evaluate our model's ability to generate try-on results that are both fidelity and realistic.

\noindent
\textbf{Implementation Details. }
The {warping module} resizes the human and garment images to $512 \times 384$ to predict the corresponding flow, which is then upsampled to the original resolution of $1024 \times 768$ before being applied to the garment image at its original size to obtain the warped garment. Using the reduced resolution for flow prediction helps minimize artifacts when applying the flow to the source garment. The {try-on module} processes images directly at their original resolution of $1024 \times 768$. The warping module is trained on two RTX A6000 GPUs for 150 epochs with a learning rate of $5 \times 10^{-5}$ and a batch size of 2 per GPU, while the try-on module is trained on four RTX A6000 GPUs for 200 epochs with a learning rate of $5 \times 10^{-4}$ and a batch size of 4 per GPU.\\

\noindent
\textbf{Baselines. } We compare our method with current state-of-the-art (SOTA) VTON models of both warping-based and warping-free approaches. For warping-based methods, we include GP-VTON~\cite{xie2023gp} and FIP-VTON~\cite{dam2024time} in our comparison. For warping-free diffusion-based methods, we compare with IDM-VTON~\cite{choi2024improving}, TPD~\cite{yang2024texture}, and CatVTON~\cite{chong2024catvton}.

\subsection{Quantitative Results}
\cref{tab:quan_results} reports a comprehensive comparison of our method against recent SOTA VTON approaches, including both warp-free and warp-based baselines. For reconstruction fidelity, our method achieves the highest PSNR (24.9) and SSIM (0.91), outperforming the best prior methods like GP-VTON (23.0 PSNR) and IDM-VTON (22.2 PSNR). In terms of perceptual quality, our approach yields the lowest FID (5.3), indicating superior realism. Additionally, our method achieves the lowest LPIPS (0.064) and DIST (0.046) scores, demonstrating significantly better perceptual similarity and structural accuracy than existing methods. Overall, these results show that our approach preserves garment details more accurately and synthesizes more realistic try-on images, advancing the state of the art.

\begin{figure*}[!ht]
    \centering
    \includegraphics[width=1\linewidth]{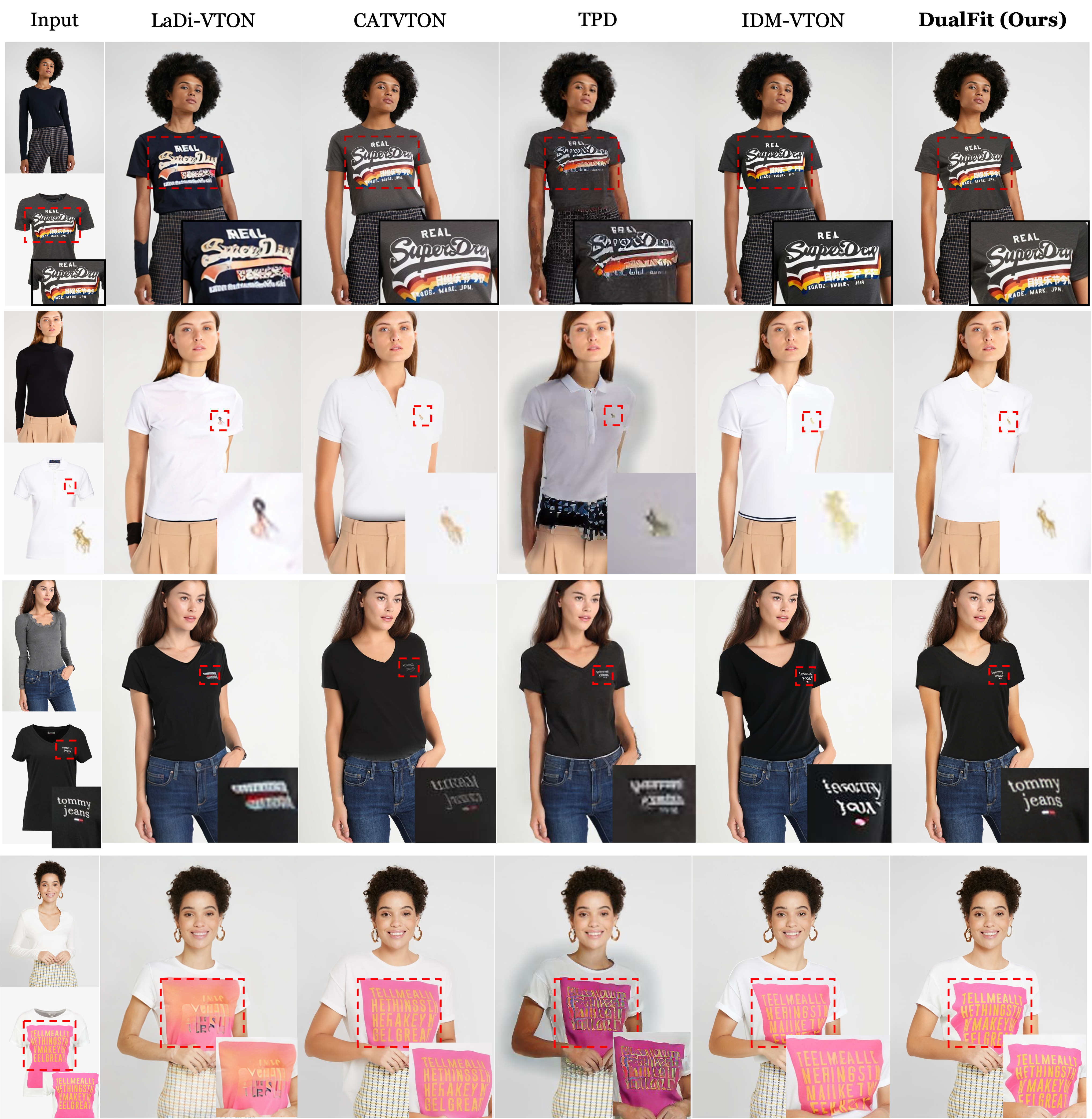}
    \caption{Qualitative comparison of our method with recent warping-free baselines based on latent diffusion models. Each row presents a distinct try-on example, with the input image on the left followed by the results from each method. Our approach consistently preserves garment textures more clearly while delivering realistic try-on images with accurate garment alignment to the target body.}

    \label{fig:quali_vs_warpfree}
\end{figure*}

\begin{figure}[!ht]
    \centering
    \includegraphics[width=1\linewidth]{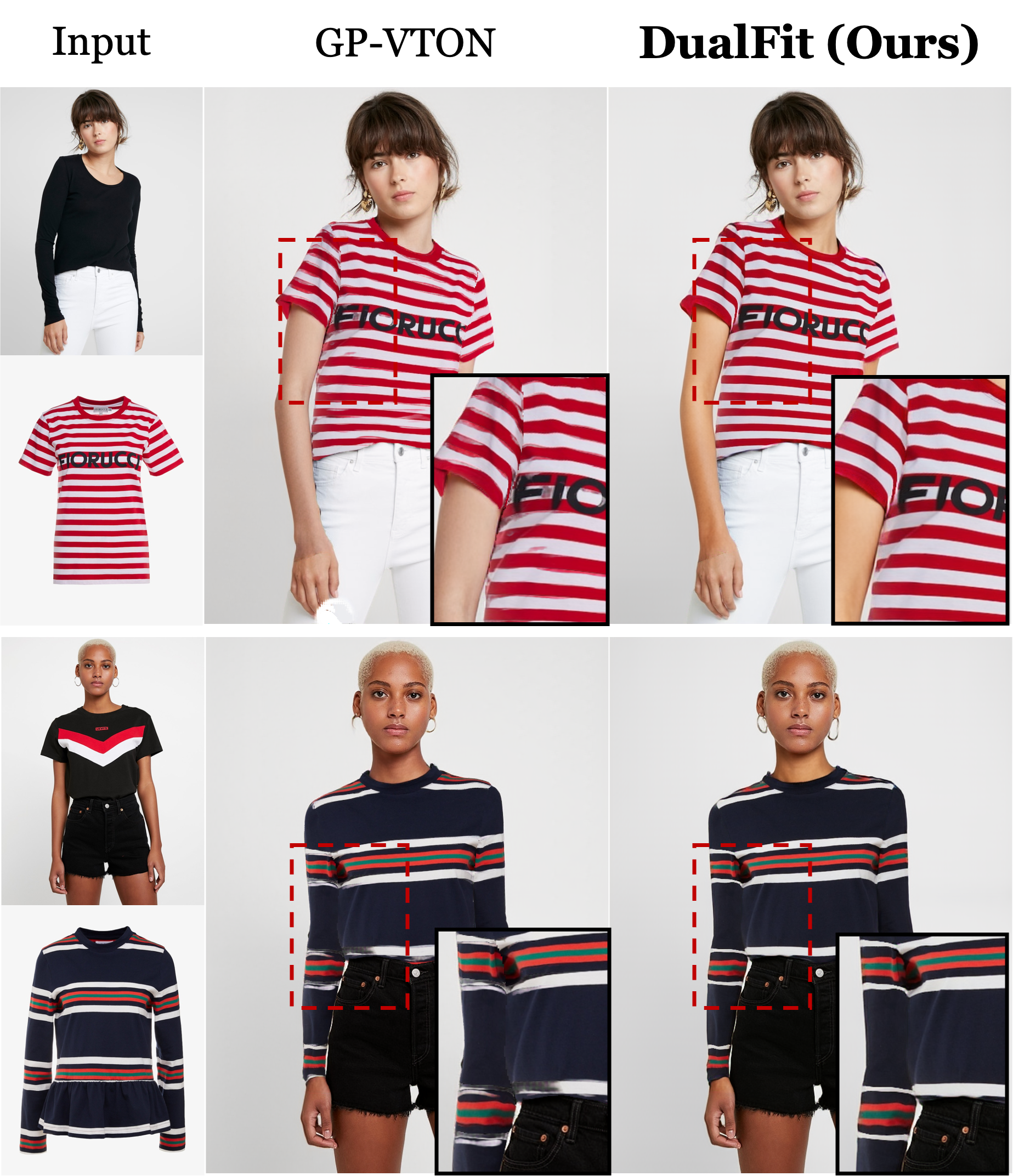}
    \caption{Qualitative comparison between our method and wapring-based method GPVTON~\cite{xie2023gp}. Best view in zoom and color.}
    \label{fig:quali_vs_gp}
\end{figure}

\subsection{Qualitative Results}
\cref{fig:quali_vs_warpfree} presents qualitative results comparing our method with recent warping-free baseline that based on latent diffusion models. Each row shows a different try-on example, with the input person and garment on the left, followed by synthesized outputs from each method. Across all examples, our approach consistently produces clearer preservation of garment textures while maintaining realistic try-on results with good garment alignment to the target body. On the other hand, diffusion-based warping-free methods all exhibits poor preservation on garment textures, graphics and printed text.

\cref{fig:quali_vs_gp} qualitatively compares the proposed method against the warping-based method GPVTON. In the first example, GPVTON result shows noticeable artifacts, particularly along the sleeve where the original long sleeve is not fully masked, and the new sleeve appears somewhat distorted. In contrast, ours demonstrates superior garment integration, with the striped t-shirt seamlessly superimposed onto the model, exhibiting more realistic drapes and a cleaner outline, as highlighted by the magnified inset of the sleeve. Similarly, the second example, GPVTON again displays visual inconsistencies, especially around the garment's edges and the transition to the model's body.  Our method, on the other hand, produces a more natural and visually coherent outcome, successfully preserving the garment's texture and shape while adapting it realistically to the model's pose, as evidenced by the detailed inset. These qualitative comparisons strongly suggest that the proposed method significantly outperforms GPVTON in generating more realistic and artifact-free VTON images.

\begin{table}
    \centering
    \caption{Ablation study on the effect of using input flat garment to condition Try-on module}
    \begin{tabular}{c|ccc}
    \toprule
        ~ & SSIM$\uparrow$ & FID$\downarrow$ & LPIPS$\downarrow$ \\ \midrule
        wo/ flat garment & 0.896 & 7.635 & 0.080 \\ 
        w/ flat garment & 0.913 & 5.261 & 0.064 \\ \bottomrule
    \end{tabular}
    \label{tab:abla_flat_garment}
\end{table}
\begin{table}
    \centering
    \caption{Ablation study on the effect of the narrow band thickness by eroding iteration}
    \begin{tabular}{c|ccc}
    \toprule
        \# iterations & SSIM$\uparrow$ & FID$\downarrow$ & LPIPS$\downarrow$ \\ \midrule
        n = 2 & 0.887 & 6.923 & 0.077 \\ 
        n = 5 & 0.913 & 5.261 & 0.064 \\
        n = 10 & 0.871 & 6.012 & 0.081 \\ \bottomrule
    \end{tabular}
    \label{tab:abla_border_thickness}
\end{table}

\subsection{Ablation Study}
\textbf{Effect of using input flat garment to condition Try-on module}. \cref{tab:abla_flat_garment} presents a comparison between conditioning the try-on module with the flat input garment and conditioning it only with the warped garment. In our method, alongside the person image $I_p$, we use the original input garment $G$ as a conditioning signal, rather than relying solely on the warped garment $G'$ as in~\cite{xie2023gp}. The results demonstrate that including the flat garment $G$ leads to better overall performance.

\textbf{Effect of the narrow band thickness by eroding iteration}. As mentioned in ~\cref{sec:tryon_module}, we obtain the narrow band of the warped garment by applying a $3 \times 3$ square kernel and performing $n$ iterations of erosion, where $n$ is a configurable parameter controlling the thickness of the narrow band. \cref{tab:abla_border_thickness} presents an ablation study on the effect of varying $n$. We observe that setting $n=5$ yields the best performance, achieving the highest SSIM of 0.913, the lowest FID of 5.261, and the lowest LPIPS of 0.064. In contrast, both smaller ($n=2$) and larger ($n=10$) erosion iterations result in worse performance across all metrics, indicating that an overly thin or overly thick narrow band degrades the quality of the try-on outputs. This highlights the importance of carefully choosing the narrow band thickness to balance detail preservation and seamless garment blending.

\section{Conclusion}
In this work, we introduced \model, a two-stage VTON pipeline designed to overcome the limitations of both warping-based and warping-free methods. While warping-free approaches offer visually smooth results, they often sacrifice high-frequency garment details due to latent-space limitations. On the other hand, traditional warping-based methods preserve garment fidelity but introduce artifacts and unnatural seams in the final outputs. \model bridges this gap by first aligning garments through a flow-based warping module and then leveraging a fidelity-preserving try-on module that selectively regenerates only necessary regions using an inpainting mask and preserved-region guidance. This design allows \model to produce try-on images that are both visually seamless and rich in fine-grained garment details, such as logos and text. Our extensive evaluations demonstrate that \model outperforms SOTA warping-free and warping-based methods across multiple metrics both reconstruction fidelity and perceptual realism. Further qualitative results confirm the effectiveness of our design, showing clearer fine-grained garment details preservation and smoother garment integration.

\noindent
\textbf{Discussion}: While \model achieves SOTA performance in both reconstruction fidelity and perceptual realism, it is not without limitations. A key dependency of our pipeline lies in the segmentation module used during the warping stage. In our current implementation, this module is trained solely on the VITON-HD dataset, which contains 11,647 training samples. Although sufficient for general upper-body try-on tasks, this dataset lacks sufficient diversity in terms of extreme body poses, occlusions, and complex real-world backgrounds. Therefore, enhancing the robustness and generalizability of the segmentation component by incorporating larger and more diverse datasets will be a promising direction for future work.

Given the efficiency of our architecture, particularly compared to computationally intensive diffusion-based methods, \model offers a promising foundation for real-time or video-based virtual try-on. As a next step, we plan to extend \model to handle temporal sequences by incorporating temporal consistency modules and designing mechanisms to stabilize garment alignment and synthesis across frames.\\

% Future works: 
% More training on segmentation for extreme cases. Moreover, with the advantages of speed, we can scale to video-based virtual try-on since other video-based virtual try-on is based on diffusion model and very computational cost. 

\noindent
\textbf{Acknowledgement.} This material is based upon work supported by the National Science Foundation (NSF) under Award No OIA-1946391 RII Track-1, Undergraduate Research Fellowship (SURF), University of Arkansas Honors College Research Grant.
{
    \small
    \bibliographystyle{ieeenat_fullname}
    \bibliography{main}
}

\end{document}